\documentclass[11pt,a4paper]{article}
\usepackage{times,latexsym}
\usepackage{url}
\usepackage[T1]{fontenc}
\usepackage{multirow}
\usepackage{color}
\usepackage{graphicx}
\usepackage{amsmath}
\usepackage{subfig}
\usepackage[table]{xcolor}
\usepackage{booktabs}

\usepackage[acceptedWithA]{tacl2021v1}

\usepackage{xspace,mfirstuc,tabulary}

\newif\iftaclinstructions
\taclinstructionsfalse 
\iftaclinstructions
\renewcommand{\confidential}{}
\renewcommand{\anonsubtext}{(No author info supplied here, for consistency with
TACL-submission anonymization requirements)}
\newcommand{\instr}
\fi

\iftaclpubformat 

\else

\fi

\title{Multi-level Shared Knowledge Guided Learning \\ for Knowledge Graph Completion}

\author{
  Yongxue Shan, Jie Zhou, Jie Peng, Xin Zhou, 
  \\
  \bf{Jiaqian Yin, Xiaodong Wang\Thanks{Xiaodong Wang is the corresponding author.}}
  \\
  \ \\
  National Key Laboratory of Parallel and Distributed Computing, College of Computer 
  \\
  National University of Denfense Technology, Changsha, China
  \\
  \texttt{\{shanyongxue001,jiezhou,pengjie,zhouxin.130,}
  \\
  \texttt{yinjiaqian,xdwang\}@nudt.edu.cn}
}

\date{}

\begin{document}
\maketitle
\begin{abstract}
In the task of Knowledge Graph Completion (KGC), the existing datasets and their inherent subtasks carry a wealth of shared knowledge that can be utilized to enhance the representation of knowledge triplets and overall performance. However, no current studies specifically address the shared knowledge within KGC. To bridge this gap, we introduce a multi-level {\bf S}hared {\bf K}nowledge {\bf G}uided learning method (SKG) that operates at both the dataset and task levels. On the dataset level, SKG-KGC broadens the original dataset by identifying shared features within entity sets via text summarization. On the task level, for the three typical KGC subtasks – head entity prediction, relation prediction, and tail entity prediction – we present an innovative multi-task learning architecture with dynamically adjusted loss weights. This approach allows the model to focus on more challenging and underperforming tasks, effectively mitigating the imbalance of knowledge sharing among subtasks. Experimental results demonstrate that SKG-KGC outperforms existing text-based methods significantly on three well-known datasets, with the most notable improvement on WN18RR (MRR: 66.6\%$\rightarrow$72.2\%, Hit@1: 58.7\%$\rightarrow$67.0\%).
\end{abstract}

\section{Introduction}
Knowledge Graphs (KGs) are directed multi-relation graphs, with entities as nodes and relations as edges, denoted as a set of triples $(h, r, t)$. Their distinctive advantage lies in efficiently representing and managing extensive knowledge, offering high-quality structured information for diverse downstream tasks, including question answering \cite{sax2020qa}, information retrieval systems \cite{bou2020ir}, and recommendation systems \cite{gao2023rs}. Despite these strengths, existing knowledge graphs still lack a substantial amount of valuable information. Effectively addressing this gap in knowledge completeness has given rise to the field of Knowledge Graph Completion (KGC). KGC aims to infer the missing entities and relations from knowledge graphs, significantly enhancing both the quality and coverage of these valuable knowledge repositories.

Existing KGC methods are mainly divided into structure-based methods and text-based methods. Structure-based methods \cite{bordes2013transe,sun2019rotate,bala2019tucker} typically map entities and relations into low-dimensional vectors and calculate the probability of valid triples by various scoring functions. Text-based methods \cite{yao2019kgbert,xie2022genkgc,kim2020mtlkgc,yao2023kgllama} adopt pre-trained language models to semantically encode textual descriptions of entities. They can encode unseen entities in training time, while making reasoning less efficient. Recent advancements, such as the bi-encoder structure proposed in studies like \citet{wang2021star, wang2022simkgc}, aim to reduce the training cost of language model encoders. This shift has led to text-based methods beginning to surpass structure-based methods in terms of performance.

\begin{figure}[ht]
	\centering
	\subfloat[Proportion of shared knowledge in triples(\%)]{\label{fig:1a-share} \includegraphics[width=.85\columnwidth]{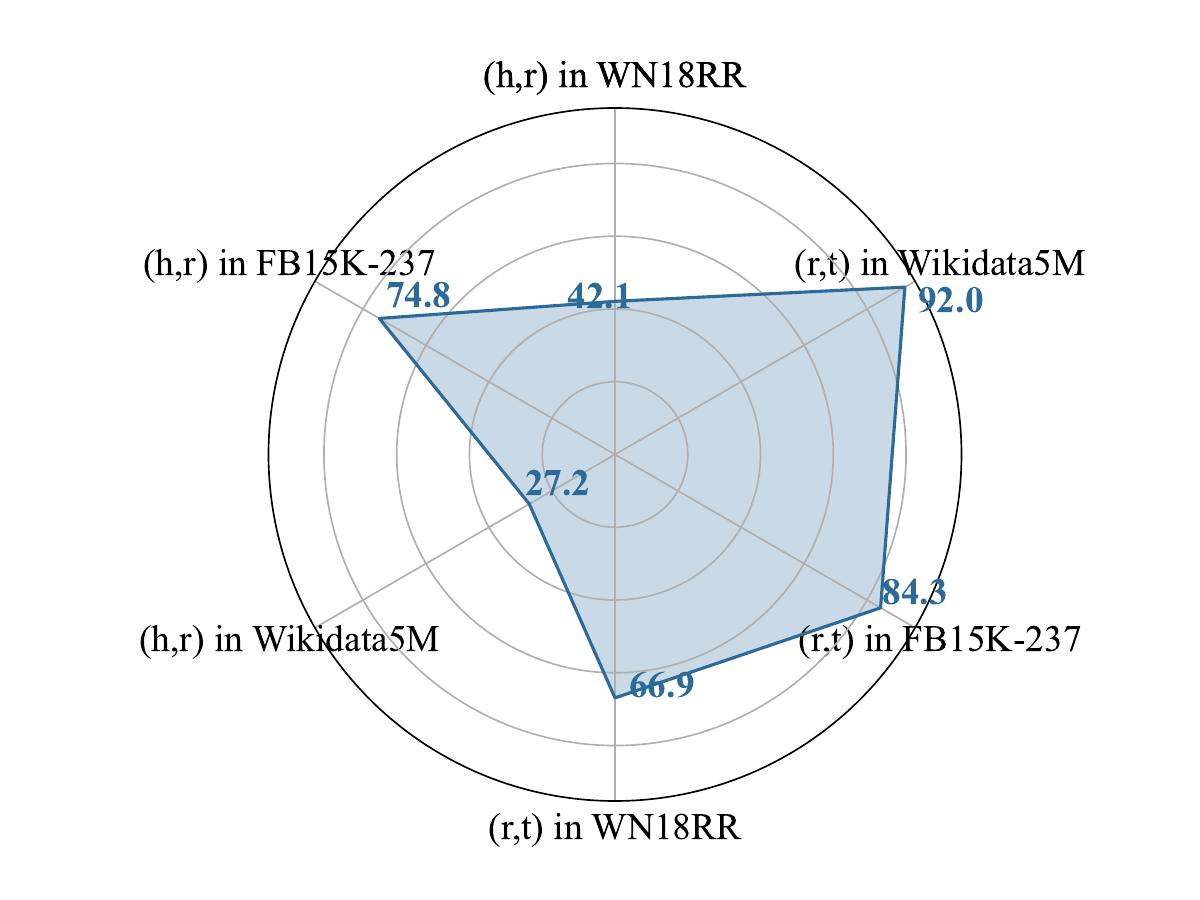}}\\
	\subfloat[Imbalance for head and tail entities]{\label{fig:1b-balance} \includegraphics[width=.85\columnwidth]{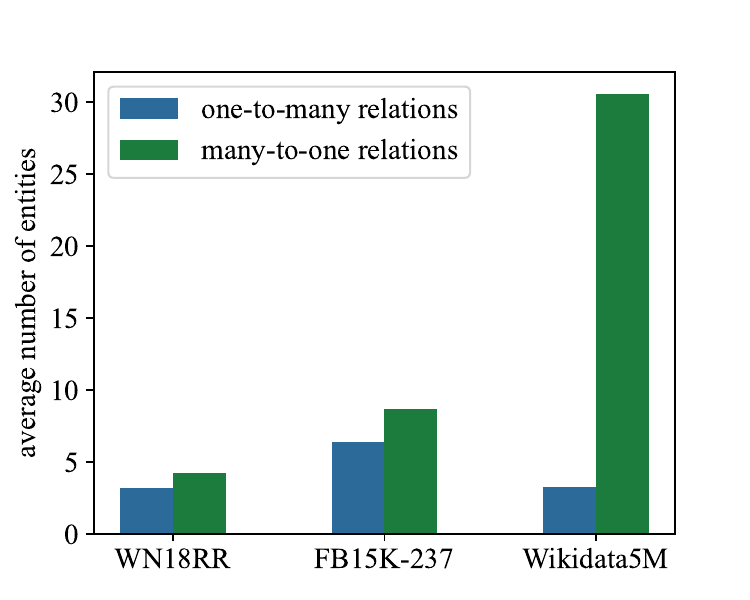}}
	\caption{(a) The greater the proportion of triples sharing the same $(r,t)$ or $(h,r)$, the more accessible knowledge we acquire. (b) The average number of connected entities in many-to-one and one-to-many relations indicates the imbalanced distribution between head entities and tail entities, respectively.}
\end{figure}

While these methods exhibit a strong capability to complete knowledge graphs, challenges persist in the effective sharing of knowledge among datasets and subtasks. Specifically, we find that the same $(h,r)$ or $(r,t)$ often appear in different triples $(h,r,t)$. According to our analysis in Figure \ref{fig:1a-share}, 42.1\% of triples can find other triples sharing the same $(h,r)$ with themselves, and 66.9\% of triples can find those sharing the same $(r,t)$ with themselves in the WN18RR training set. For instance, \textit{(Kirsten Dunst, film actor, Spider-Man), (Willem Dafoe, film actor, Spider-Man), (James Franco, film actor, Spider-Man)} all have the same relations and tail entities. This suggests the potential existence of shared knowledge, such as "\textit{American film actor}," among various head entities. Leveraging this dataset-level shared knowledge is essential to enhance the learning ability of triples and assist the model in correctly identifying answers from lexically similar candidates.

Notably, we observe considerable performance variations across various KGC subtasks, even when applied to the same dataset. For KG-BERT \cite{yao2019kgbert} on the WN18RR dataset, the Hit@10 prediction results differ notably for head entities (54\%) and tail entities (60.7\%). This discrepancy arises from certain relations, such as gender and city, linking more head (tail) entities and fewer tail (head) entities. As shown in Figure \ref{fig:1b-balance}, the issue of imbalanced distribution of head entities and tail entities is prevalent in knowledge graphs, yet it receives limited attention in research. Existing multi-task learning methods treat head entity and tail entity prediction equally, ignoring the intricacies of more complex tasks. Therefore, making the model focus on more challenging tasks while learning the shared knowledge across multiple subtasks becomes an urgent concern. This task-level shared knowledge can enhance the model's learning of entity and relation embeddings. 

In this paper, we introduce a multi-level {\bf S}hared {\bf K}nowledge {\bf G}uided learning method ({\bf SKG}) for knowledge graph completion. To capture dataset-level shared knowledge within specific entity sets, we jointly train original triples, triples with identical head entities and relations, and triples with identical relations and tail entities. For task-level knowledge sharing, we incorporate relation prediction in multi-task learning to assist entity prediction task, enabling the model to acquire more relation-aware entity information. In each iteration, our loss weight allocation scheme assigns higher loss weights to tasks that are more challenging and underperforming, effectively addressing the imbalanced distribution of head and tail entities. In summary, our contributions include:
\begin{itemize}
\item We extract dataset-level shared knowledge by extending the original dataset, bolstering the model's ability to identify correct answers from lexically similar candidates in the bi-encoder architecture.
\item We design a novel multi-task learning architecture with dynamically adjusted loss weights for task-level knowledge sharing. This ensures the model focuses more on challenging and underperforming tasks, alleviating the imbalance of subtasks in KGC.
\item SKG-KGC is evaluated on three benchmark datasets: WN18RR, FB15k-237 and Wikidata5M. Experimental results demonstrate the competitive performance of our model in both transductive and inductive settings, with notable success on the WN18RR dataset.

\end{itemize}

\section{Related Work}
{\bf Knowledge graph completion} KGC has been extensively studied for many years as a popular research topic. It can be divided into three subtasks: head entity prediction, relation prediction and tail entity prediction. Structure-based methods, such as TransE \cite{dou2021transmtl}, RotatE \cite{sun2019rotate}, TuckER \cite{bala2019tucker} and Complex-N3 \cite{jain2020ComplexN3}, map entities and relations to low-dimensional vector spaces and measure the plausibility of triples by various scoring functions. Recent text-based methods represented by KG-BERT \cite{yao2019kgbert} attempt to integrate pre-trained language models for encoding textual descriptions of entities and relations. PKGC \cite{lv2022pkgc} converts each triple into natural prompt sentences, utilizing a single encoder for triple encoding. \citet{xie2022genkgc}, \citet{sax2022kgt5}, \citet{yao2023kgllama} formulate KGC as a sequence-to-sequence generation task and explore Seq2Seq PLM models to directly generate required text. StAR \cite{wang2021star} simultaneously learns graph embeddings and contextual information of the text encoding method. \citet{chen2023csprom} employs conditional soft prompts to integrate textual description and structural knowledge. In contrast, SimKGC \cite{wang2022simkgc} introduces contrastive learning and a bi-encoder with a pre-trained language model to encode entities and relations separately. It proves highly efficient for training with a large negative sample size, enhancing the efficiency of KGC training and inference.

{\bf Multi-task Learning} MTL aims to concurrently train deep learning models by leveraging information from multiple interconnected tasks. Balancing losses during training facilitates tasks in providing valuable insights to each other, resulting in a more proficient and robust model. For the KGC task, \citet{kim2020mtlkgc} first propose a multi-task learning method, integrating relation prediction, relevance ranking, and link prediction tasks. Subsequent models focus on introducing additional knowledge or potent pre-trained language models (PLM). For instance, \citet{dou2021transmtl} propose a novel embedding framework for multi-task learning, enabling the transfer of structural knowledge across different KGs. Incorporating the ALBERT-large \cite{lan2020albert} model with more parameters as the text encoder, \citet{tian2022mitkgc} enhances model performance at the expense of increased training costs. Meanwhile, \citet{li2023lpbert} employ a multi-task pre-training strategy to capture relational information and unstructured semantic knowledge within structured knowledge graphs. These studies emphasize the interconnectedness of various KGC subtasks, highlighting that knowledge sharing among them can enhance overall performance.

However, they overlook the distinction between head entity prediction and tail entity prediction tasks, which arises from the imbalanced distribution of head and tail entities. Recognizing this, our SKG-KGC model explicitly distinguishes between head entity prediction and tail entity prediction in the context of multi-task learning. We attempt to achieve superior performance and scalability by employing the basic PLM model and fewer subtasks.

\section{Method}
In this section, we introduce a multi-level {\bf S}hared {\bf K}nowledge {\bf G}uided learning method ({\bf SKG}) for knowledge graph completion. We elaborate the entire architecture of the proposed model in section \ref{sec:Model Structure}. In sections \ref{sec:Input Layer} and \ref{sec:Multi-Task Learning}, we illustrate how our method captures shared knowledge at both dataset and task levels for KGC. These insights are seamlessly integrated at the bi-encoder architecture, as explained in section \ref{sec:Encoding Layer}. The following sections provide a detailed overview of the training and inference processes of our model.

\subsection{Model Structure}
\label{sec:Model Structure}

\begin{figure*}[ht]
\centering
\includegraphics[scale=0.5]{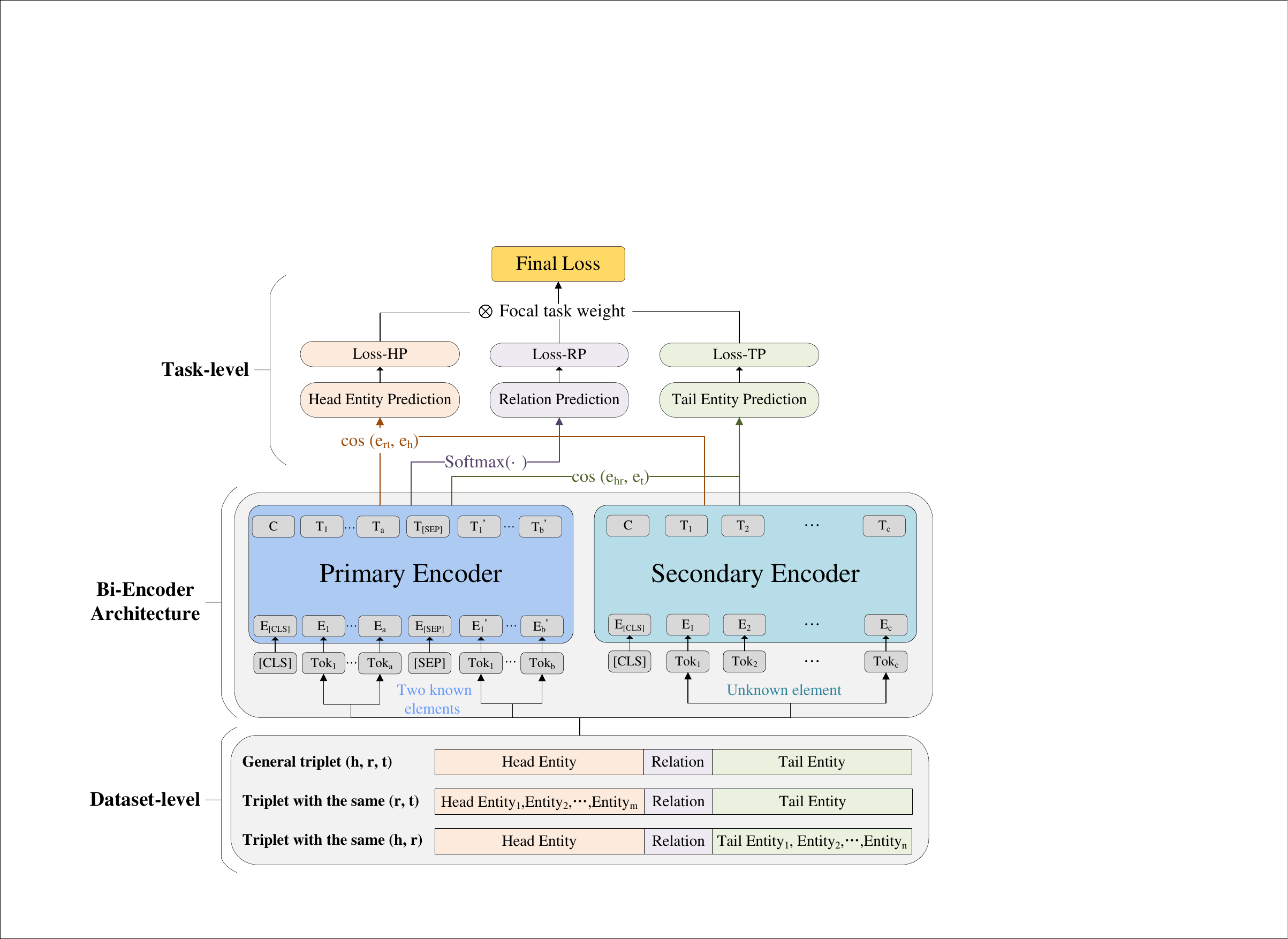}
\caption{\label{fig:2-Model}An overview of the SKG-KGC model.}
\end{figure*}

Figure \ref{fig:2-Model} illustrates the overview of the SKG-KGC model. Our model consists of three parts:
\begin{itemize}
\item Dataset level: During training, the model is simultaneously trained with original triples, triples with identical $(h,r)$, and triples with identical $(r,t)$. This approach strengthens the learning of shared features among entity sets while reducing text redundancy.
\item Bi-Encoder Architecture: Two encoders are initialized with the same pre-trained model but do not share parameters. The primary encoder computes the joint embedding of the two known elements in triples, while the secondary encoder computes the representation of the missing entities.
\item Task level: We design balanced multi-task learning by introducing a relation prediction subtask to assist link prediction. In each iteration, the model assigns higher loss weights to challenging and underperforming subtasks, facilitating dynamic knowledge sharing across different subtasks.
\end{itemize}

\subsection{Dataset Expansion}
\label{sec:Input Layer}
In addition to the original triples $(h,r,t)$, our proposed model also incorporates triples with the same head entity and relation $(h,r,\{t_0,t_1,...,t_i\})$ and triples with the same relation and tail entity $(\{h_0,h_1,...,h_j\},r,t)$. Common features among different entities in triples are identified through text summarization. For instance, $(h_1,r,t)$, $(h_2,r,t)$ and $(h_3,r,t)$ are valid triples in the training dataset, where the relation $r$ and the tail entity $t$ are consistent. Consequently, the three head entities $h_1,h_2,h_3$ may share common or similar features. If a new entity $h_0$ also contains these common features within the head entity set $\{h_1, h_2, h_3\}$, the triplet $(h_0,r,t)$ is more likely to be considered reasonable.

The model takes text sequences as input, corresponding to the three types of triples for knowledge graph completion. Each entity text sequence comprises the entity's name and its corresponding text description. For the triple (\{04692908, 00387897\}, derivationally related form, 01259005), the input sequence is: "\emph{[CLS] \textbf{chip}, a mark left after a small piece has been chopped or broken off of something [PSEP] \textbf{snick}, a small cut [SEP] derivationally related form [SEP] nick, cut a nick into [SEP]}". The bold font indicates the name of each head entity. [PSEP] serves as the separator for entities in the head entity set. The use of [CLS] and [SEP] aligns with the BERT-base model. Further details regarding different subtasks are provided in Table \ref{tab:subtask example}.

\begin{table*}[ht]
\centering
\begin{tabular}{c|ccc|cc}
\hline
\multirow{2}{*}{\bf Subtask} & \multicolumn{3}{c|}{\bf Input} & \multirow{2}{*}{\bf Label} & \multirow{2}{*}{\bf Type} \\
& \bf h tokens & \bf r tokens & \bf t tokens &  &  \\ \hline
HP & [MASK] & has part & China & Asia & Candidate entity ranking \\ 
RP & Asia & [MASK] & China & has part & Multi-classification \\ 
TP & Asia & has part & [MASK] & China & Candidate entity ranking \\ \hline
\end{tabular}
\caption{\label{tab:subtask example} An example of the head entity prediction (HP), relation prediction (RP) and tail entity prediction (TP) subtasks in KGC.}
\end{table*}

However, the integration of entity sets may result in more extensive and redundant textual information. To address this, we utilize the TextRank algorithm \cite{mi2004textrank} to refine the input sequence from entity sets. Initially, we construct an undirected graph by segmenting the text into $m$ sentence units. Each sentence unit is treated as a node in the graph. We then quantify the similarity between sentences using a word overlap-based method. This similarity score $w$ directly influences the weight of the graph's edges, with higher similarity resulting in greater edge weight. Next, we assign an initial TextRank value of $1/m$ to each sentence unit. These values are iteratively updated according to the following formula:
\begin{equation}
T(S_i)=(1-d)+d*\sum_{S_j}\frac{w_{ji}}{\sum_{S_k}w_{jk}}T(S_j)
\end{equation}

Here, $S_j$ and $S_k$ are the nodes pointing to and pointed by $S_i$ respectively, $T(S_j)$ denotes the TextRank value of the $j$-th sentence, $w_{ji}$ and $w_{jk}$ are the weight of edges between nodes (sentence similarity), and $d$ is the damping ratio, signifying the probability of jumping from one node to another. 

After iteration, we obtain the final TextRank value $T(S_i)$ for the $i$-th sentence unit. The top-$n$ sentences with the highest TextRank values are then selected as concise text, providing the model with essential yet condensed descriptive information.

\subsection{Bi-Encoder Architecture}
\label{sec:Encoding  Layer}
Unlike MTL-KGC \cite{kim2020mtlkgc} using a single encoder, our proposed model employs two encoders initialized with the same pre-trained language model but without sharing parameters. Each encoder autonomously acquires shared knowledge at both the dataset and task levels. The primary encoder computes the joint embedding of the two known elements in triples, while the secondary encoder computes the representation of the missing entities.

{\bf Head Entity Prediction} Given a triple $(h,r,t)$, the main encoder concatenates the text descriptions of the relation $r$ and the tail entity $t$ with [SEP], calculating the relationship-aware embedding of the tail entity $(r, t)$. Meanwhile, the secondary encoder encodes the head entity $h$. This method, as demonstrated in previous approaches \cite{wang2021star, wang2022simkgc}, proves to be more efficient than encoding the entire triple simultaneously. Subsequently, we employ mean pooling and L2 normalization strategies to derive fixed-size embeddings. The similarity score between the head entity embedding $e_h$ and the relation-aware tail entity embedding $e_{rt}$ is determined by cosine similarity, expressed by the following formula:
\begin{equation}
f_\text{HP}(h,r,t)=\cos \left(e_h, e_{rt}\right)=\frac{e_h \cdot e_{rt}}{\left\|e_h\right\|\left\|e_{rt}\right\|}
\end{equation}

{\bf Tail Entity Prediction} The identical bi-encoder computes the similarity score between the tail entity embedding $e_t$ and the relation-aware head entity embedding $e_{hr}$ for tail entity prediction $(h,r,?)$. The formula is as follows:
\begin{equation}
f_\text{TP}(h,r,t)=\cos \left(e_t, e_{hr}\right)=\frac{e_t \cdot e_{hr}}{\left\|e_t\right\|\left\|e_{hr}\right\|}
\end{equation}

We incorporate the idea of contrastive learning to make the anchor point closer to positive samples $(h, r, t)$ and farther from negative samples $(h^{\prime}, r, t)$ or $(h, r, t^{\prime})$. The proper selection of negative samples significantly impacts the training model's performance. For ease of comparison, our model employs negative samples consistent with those constructed in SimKGC.

{\bf Relation Prediction} The goal of relation prediction $(h,?,t)$ is to predict the missing relation between two given entities. Due to the absence of detailed descriptions of relations, separately encoding relations, as done in the previous two tasks, is impractical. Therefore, we treat relation prediction as a multi-classification task. The scoring function $g$ for the relation label is expressed as follows:
\begin{equation}
g(h,r,t)=\text{softmax}(e_{ht} W_{RP}^T)
\end{equation}

Here, $e_{ht}$ represents the head entity and tail entity embeddings encoded by the shared main encoder, and $W_{RP}$ is the parameter matrix of the classification layer used for relation prediction.

\subsection{Balanced Multi-Task Learning}
\label{sec:Multi-Task Learning}
In multi-task learning, simultaneously training multiple tasks can be challenging or inefficient without achieving a proper balance among them. Thus, we introduce a dynamic and balanced multi-task weight allocation scheme to ensure equilibrium among the three subtasks: head entity prediction, relation prediction, and tail entity prediction. This approach dynamically takes into account the learning difficulty and accuracy of tasks during each iteration, assigning a higher loss weight to tasks that are challenging to learn and exhibit lower performance. We use $d_k(t)$ to assess the difficulty and accuracy of task $k$ in the $t$-th epoch as follows:
\begin{equation}
d_k(t)=-\left(1-\bar{a}_k(t-1)\right)^{r_k} \log \left(\bar{a}_k(t-1)\right)
\end{equation}

Here, $d_k(t)$ is calculated similarly to focal loss \cite{lin2020focal}, augmenting the weight of difficult-to-distinguish samples. Although focal loss is originally designed for classification \cite{rom2020electrocardiogram}, we extend its application to multi-task weight assignment. For task $k$, $\bar{a}_k(t-1) \in(0,1)$ denotes the normalized accuracy metric of the validation set during the iteration immediately before $t$. An increased accuracy metric $\bar{a}_k(t)$ indicates enhanced learning capability of the model for the task, thus suggesting a reduction in weight allocation. The focusing parameter $r_k$ smoothly adjusts the proportion of tasks that are down-weighted. As the task becomes simpler, it is accorded less weight.

In this paper, the focusing parameter $r_k$ primarily mirrors the learning difficulty of the head entity prediction and tail entity prediction tasks, denoted as the ratio of the average number of connected entities in many-to-one and one-to-many relations. A higher count of entities connected by many-to-one relations increases the learning complexity of the head entity prediction task. The tail entity prediction task is also influenced by the number of entities connected by one-to-many relations. The default value of $r_k$ for the relation prediction task is set to 1.

Subsequently, we normalize $d_k(t)$ using the softmax function and multiply it by the number of tasks $K$, ensuring $K=\sum_i w_i(t)$. Finally, we obtain the loss weight $w_k(t)$ for task $k$ in the $t$-th epoch.
\begin{equation}
w_k(t)=\frac{K \exp \left(d_k(t)\right)}{\sum_i \exp \left(d_i(t)\right)}
\end{equation}

For $t$ =1, we initialize the loss weight $w_k(t)$ of each task to 1, though introducing any non-balanced initialization weight based on prior knowledge is also viable.

\subsection{Training}
\label{sec:Training}
For different subtasks in KGC, we optimize our proposed model using InfoNCE loss and cross-entropy loss, respectively. 

{\bf InfoNCE loss} We treat both head entity prediction (HP) and tail entity prediction (TP) as candidate entity ranking tasks. Therefore, we employ the InfoNCE loss with additive margin softmax \cite{yang2019BiDE,wang2022simkgc} for $\mathcal{L}_{\text{HP}}$ and $\mathcal{L}_{\text{TP}}$. The loss $\mathcal{L}_{\text{TP}}$ is defined as follows:
\begin{equation}
\mathcal{L}_{\text{TP}}=-\log \frac{e^{(f(h, r, t)-\gamma) / \tau}}{e^{(f(h, r, t)-\gamma) / \tau}+\sum_{i=1}^{|\mathcal{N}|} e^{f\left(h, r, t_i^{\prime}\right) / \tau}}
\end{equation}

The scoring function $f(h, r, t) \in[-1,1]$ for triples is the cosine similarity of $\mathbf{e}_{hr}$and $\mathbf{e}_{t}$. The additive margin $\gamma>0$ enhances the separation between true triples and false triples. We utilize the temperature $\tau$ to adjust the relative importance of negatives in triples and introduce $\log \frac{1}{\tau}$ as a learnable parameter during training. $|\mathcal{N}|$ represents the number of negative samples. The same approach is applicable to obtain the loss $\mathcal{L}_{\text{HP}}$.

{\bf Cross-entropy loss} For the relation prediction subtask (RP), we train the model by minimizing the cross-entropy loss between the true relations and the predicted relations. The cross-entropy loss function is given by:
\begin{equation}
\mathcal{L}_{\text{RP}}=-\sum_{(h,r,t)\in D_{\text{RP}}}y \log g(h,r,t)
\end{equation}
Here, $y$ is a one-hot encoding of relation $r$. $g(h,r,t)$ denotes the output score of the relation prediction. 

Throughout the training process, we employ the mini-batch stochastic gradient descent algorithm to optimize the objective function. For each training step, a mini-batch is randomly selected from the entire training dataset $D=D_{\text{HP}}\cup D_{\text{RP}}\cup D_{\text{TP}}$. Subsequently, the model is trained sequentially for the task associated with the specific mini-batch. Finally, we multiply the three loss functions by their corresponding weights and sum them up to obtain the overall loss function $\mathcal{L}$. The formula is as follows:
\begin{equation}
\mathcal{L}=\mathcal{L}_{\text{HP}} \cdot w_{\text{HP}}+\mathcal{L}_{\text{RP}} \cdot w_{\text{RP}}+\mathcal{L}_{\text{TP}} \cdot w_{\text{TP}}
\end{equation}

\subsection{Inference}
\label{sec:Inference}
Assume there are $|T|$ test triples and $|E|$ candidate entities in the head entity prediction task. Traditional cross encoders, such as KG-BERT \cite{yao2019kgbert} and MTL-KGC \cite{kim2020mtlkgc}, traverse $|E|$ entities for each test triple $(?,r,t)$. They replace the head entity in the test triplet repeatedly and select the highest-ranking entity as the candidate. This means a test triple requires $|E|$ computations, and $|T|$ triples need $|E|\times|T|$ computations in total. In contrast, our method employs two independent encoders similar to SimKGC \cite{wang2022simkgc}. The primary encoder computes the relation-aware tail entity embeddings for $|T|$ test triples, while the secondary encoder necessitates only a one-time computation for $|E|$ candidate entities without re-traversing all entities. The embeddings from the two encoders are combined using a dot product operation to obtain the ranking scores for all entities. This reduces the required BERT forward passes to $|E|+|T|$, significantly reducing inference time. 

Likewise, the reasoning process for the tail entity prediction subtask follows a comparable pattern. The computational complexity also shifts from $|E|\times|T|$ to $|E|+|T|$. The inference complexity of the relation prediction subtask remains $|T|$, owing to the retention of the cross-encoder. Moreover, we have the capability to pre-compute the embeddings of unseen entities or relations based on their text descriptions. Consequently, our model can also facilitate inductive reasoning for some unseen entities or relations.

\section{Experiments}
\subsection{Experimental Setup}

{\bf Dataset} Our model is evaluated on three benchmark datasets: WN18RR \cite{Det2018wn18rr}, FB15k-237 \cite{tou2015fb15k237}, and Wikidata5M \cite{wang2021kepler}. Further details regarding dataset statistics are provided in Table \ref{tab:dataset}. WN18RR is a subset of WordNet \cite{Miller1995wordnet}, containing about 41k entities and 11 semantic relations between words. FB15k-237, a subset of FreeBase \cite{bol2008freebase}, consists of about 15k entities and 237 relations. For text descriptions in WN18RR and FB15k237, we follow the data provided by KG-BERT \cite{yao2019kgbert}. Wikidata5M integrates the Wikidata knowledge graph and Wikipediapages, comprising nearly 5 million entities and about 20 million triples. It is used for both transductive and inductive KGC tasks. In the transductive setting, entities appearing in the test set are encountered in the training set, while in the inductive setting, entities in the test set have never appeared in the training set.

\begin{table*}[ht]
\centering
\begin{tabular}{c|ccccc}
\hline
\bf Dataset & \bf \#entity & \bf \#relation & \bf \#train & \bf \#valid & \bf \#test\\ \hline
WN18RR & 40,943 & 11 & 86,835 & 3,034 & 3,134  \\
FB15K-237 & 14,541 & 237 & 272,115 & 17,535 & 20,466  \\ 
Wikidata5M-Trans & 4,594,485 & 822 & 20,614,279 & 5,163 & 5,163  \\
Wikidata5M-Ind & 4,579,609 & 822 & 20,496,514 & 6,699 & 6,894 \\ \hline
\end{tabular}
\caption{\label{tab:dataset} Statistics of the datasets used in this paper. “Wikidata5M-Trans” and “Wikidata5M-Ind” refer to the transductive and inductive settings, respectively.}
\end{table*}

{\bf Evaluation Metrics} For each test triple $(h, r, t)$, our model predicts the tail entity $t$ by ranking all entities based on $(h,r)$, and similarly, predicts the head entity $h$ by ranking all entities based on $(r,t)$. The evaluation employs four metrics: mean reciprocal rank (MRR), Hit@1, Hit@3 and Hit@10. MRR is the average reciprocal rank of all test triples, while Hit@k represents the proportion of correct entities ranked within the top-k candidates. All metrics are reported under the filtered setting \cite{bordes2013transe}, and computations involve averaging over head entity prediction $(?,r,t)$ and tail entity prediction $(h,r,?)$ tasks.

{\bf Hyperparameters} The SimKGC model \cite{wang2022simkgc} serves as our benchmark, with most hyperparameters aligning with it. The encoders are initialized with \emph{BERT-base-uncased} (English). The AdamW optimizer with linear learning rate decay is employed. All models are trained with batch size 1024 on 4 A100 GPUs. We conduct a grid search on learning rates within $\{10^{-5}, 3\times10^{-5}, 5\times10^{-5}\}$. Entity descriptions are truncated to a maximum of 50 tokens. In the TextRank algorithm, we set the damping ratio $d$ at 0.85 and select the top three sentences as the summarized text. Each task's initial weight in multitask learning is set to 1. The temperature $\tau$ initializes at 0.05, and the additive margin $\gamma$ for InfoNCE loss is 0.02. For the WN18RR, FB15k-237, and Wikidata5M datasets, we train for 50, 10, and 1 epochs, respectively.

\subsection{Main Results}
We compare the performance of SKG-KGC with state-of-the-art baseline models, covering both structure-based methods and text-based methods. Table \ref{tab:mainwnfb} illustrates the main results on the WN18RR and FB15K-237 datasets, while Table \ref{tab:mainwiki} shows the performance on the Wikidata5M dataset under transductive and inductive settings.

\begin{table*}[ht]
\centering
\resizebox{\linewidth}{!}{
\begin{tabular}{c|cccc|cccc}
\hline
\multirow{2}{*}{\bf Model} & \multicolumn{4}{c|}{\bf WN18RR} & \multicolumn{4}{c}{\bf FB15K-237} \\
& \bf MRR & \bf Hit@1 & \bf Hit@3 & \bf Hit@10 & \bf MRR & \bf Hit@1 & \bf Hit@3 & \bf Hit@10 \\ \hline
\multicolumn{9}{l}{\it Structure-based Methods} \\ \hline
TransE \cite{bordes2013transe} \dag & 24.3 & 4.3 & 44.1 & 53.2 & 27.9 & 19.8 & 37.6 & 44.1 \\
ComplEx \cite{tro2016complex} \dag& 44.9 & 40.9 & 46.9 & 53.0 & 27.8 & 19.4 & 29.7 & 45.0 \\
RotatE \cite{sun2019rotate} \dag & 47.6 & 42.8 & 49.2 & 57.1 & 33.8 & 24.1 & 37.5 & 53.3 \\
TuckER \cite{bala2019tucker} \dag & 47.0 & 44.3 & 48.2 & 52.6 & 35.8 & 26.6 & 39.4 & 54.4 \\
Complex-N3 \cite{jain2020ComplexN3} & 49.0 & 44.0 & - & 58.0 & \bf{37.0} & \bf{27.0} & - & 56.0 \\
TransMTL-H \cite{dou2021transmtl} & 49.8 & - & - & 57.0 & 34.9 & - & - & 53.7 \\
SEPA \cite{gre2023sepa}& 48.1 & 44.1 & 49.6 & 56.2 & 33.2 & 24.3 & 36.3 & 50.9 \\ \hline
\multicolumn{9}{l}{\it Text-based Methods} \\ \hline
KG-BERT \cite{yao2019kgbert} \dag & 21.6 & 4.1 & 30.2 & 52.4 & - & - & - & 42.0 \\
MTL-KGC \cite{kim2020mtlkgc} & 33.1 & 20.3 & 38.3 & 59.7 & 26.7 & 17.2 & 29.8 & 45.8 \\
StAR \cite{wang2021star} & 40.1 & 24.3 & 49.1 & 70.9 & 29.6 & 20.5 & 32.2 & 48.2 \\
GenKGC \cite{xie2022genkgc} & - & 28.7 & 40.3 & 53.5 & - & 19.2 & 35.5 & 43.9 \\
MIT-KGC \cite{tian2022mitkgc} & - & 33.5 & 58.2 & 76.5 & - & 21.2 & \bf{41.7} & \bf{57.5} \\
SimKGC \cite{wang2022simkgc} & 66.6 & 58.7 & 71.7 & 80.0 & 33.6 & 25.7 & 37.3 & 49.8 \\ 
LP-BERT \cite{li2023lpbert} & 48.2 & 34.3 & 56.3 & 75.2 & 31.0 & 22.3 & 33.6 & 49.0 \\ 
SKG-KGC & \bf{72.2} & \bf{67.0} & \bf{75.1} & \bf{81.6} & 35.0 & 26.4 & 37.7 & 52.2 \\ \hline
\end{tabular}}
\caption{\label{tab:mainwnfb} Main results for WN18RR and FB15K-237 datasets. Results of [\dag] are from StAR \cite{wang2021star} and the other results are from the corresponding papers. Bold numbers represent the best results.}
\end{table*}

\begin{table*}[ht]
\centering
\resizebox{\linewidth}{!}{
\begin{tabular}{c|cccc|cccc}
\hline
\multirow{2}{*}{\bf Model} & \multicolumn{4}{c|}{\bf Wikidata5M-Trans} & \multicolumn{4}{c}{\bf Wikidata5M-Ind} \\
& \bf MRR & \bf Hit@1 & \bf Hit@3 & \bf Hit@10 & \bf MRR & \bf Hit@1 & \bf Hit@3 & \bf Hit@10 \\ \hline
\multicolumn{9}{l}{\it Structure-based Methods} \\ \hline
TransE \cite{bordes2013transe} \ddag & 25.3 & 17.0 & 31.1 & 39.2 & - & - & - & - \\
RotatE \cite{sun2019rotate} \ddag & 29.0 & 23.4 & 32.2 & 39.0 & - & - & - & - \\ \hline
\multicolumn{9}{l}{\it Text-based Methods} \\ \hline
DKRL \cite{Xie2016dkrl} \ddag & 16.0 & 12.0 & 18.1 & 22.9 & 23.1 & 5.9 & 32.0 & 54.6 \\
KEPLER \cite{wang2021kepler} \ddag & 21.0 & 17.3 & 22.4 & 27.7 & 40.2 & 22.2 & 51.4 & 73.0 \\
BLP-SimplE \cite{daza2021blp} \ddag & - & - & - & - & 49.3 & 28.9 & 63.9 & 86.6 \\
SimKGC \cite{wang2022simkgc} & 35.8 & 31.3 & 37.6 & 44.1 & 71.4 & 60.9 & 78.5 & \bf{91.7} \\
KGT5 \cite{sax2022kgt5} & 30.0 & 26.7 & 31.8 & 36.5 & - & - & - & - \\
SKG-KGC & \bf{36.6} & \bf{32.3} & \bf{38.2} & \bf{44.6} & \bf{72.0} & \bf{61.6} & \bf{78.8} & \bf{91.7} \\ \hline
\end{tabular}}
\caption{\label{tab:mainwiki} Main results for Wikidata5M dataset. Results of [\ddag] are from SimKGC \cite{wang2022simkgc} and the other results are from the corresponding papers. We follow the evaluation protocol used in SimKGC.}
\end{table*}

On the WN18RR dataset, the SKG-KGC model outperforms other models significantly. It exhibits notable improvements over the state-of-the-art (SOTA) method in MRR, Hit@1, Hit@3, and Hit@10, with gains of 5.6\%, 8.3\%, 3.4\%, and 1.6\% respectively. The most substantial enhancement is observed in Hit@1, potentially attributed to the presence of more lexically similar entities and a sparser graph structure in the WN18RR dataset. We argue that shared knowledge aids the model in learning crucial textual descriptions, enhancing its ability to identify similar candidate entities. The dynamic and balanced loss weight scheme in multi-task learning enables the model to concentrate more on specific subtasks, enhancing its efficacy in handling sparse data in WN18RR. Moreover, text-based methods consistently outperform structure-based methods, underscoring their advantage in grasping the semantics of words.

Compared to the WN18RR dataset, the FB15K-237 dataset features richer relations and fewer entities. Our model exhibits improved experimental performance among text-based methods, with the exception of MIL-KGC, which utilizes the more potent AlBERT-large encoder and undergoes longer training times. This outcome underscores the effectiveness of shared knowledge and balanced multi-task learning in SKG-KGC for leveraging text information. However, our model still falls short when compared to structured methods like TuckeER and Complex-N3. Two main reasons contribute to this shortfall. Firstly, the limited number of entities in the FB15K-237 dataset results in inadequate learning of entity textual descriptions. Additionally, structured methods contribute to a more effective understanding of generalizable inference rules, which proves advantageous for the FB15K-237 dataset.

The Wikidata5M dataset spans various domains and boasts a much larger scale compared to WN18RR and FB15K-237. As indicated in Table \ref{tab:mainwiki}, our model demonstrates state-of-the-art (SOTA) performance in both transductive and inductive settings when compared to existing structure-based and text-based methods. Notably, the million-scale data results in a prolonged training time for our model in a single iteration. To facilitate comparisons and minimize training costs, we adopt the approach from SimKGC, maintaining the epoch at 1 during training. Consequently, the dynamic and balanced loss weight allocation scheme is not applied to this dataset. Although extending the existing dataset and incorporating the relation prediction subtask in multi-task learning contribute to some performance enhancement, further improvements can be achieved. Additionally, the exceptional performance on the Wikidata5M\_inds dataset underscores our model's capability to infer entities not encountered in the training set.

\subsection{Ablation Studies}
We conduct the ablation studies to explore the impact of each specific component on the SKG-KGC model. Specifically, "w/o dataset expansion" means that the model is trained only using original triples. "w/o balanced multi-task learning" refers to treating the loss weights of multiple subtasks as 1. "w/o multi-level shared knowledge" means removing both components that gather dataset-level and task-level knowledge. "w/o bi-encoder architecture" indicates that we only use one encoder for all triple elements. The results shown in Table \ref{tab:ablation} highlight that removing any of these components greatly reduces the model's performance.

\begin{table*}[ht]
\centering
\resizebox{\linewidth}{!}{
\begin{tabular}{c|cccc|cccc}
\hline
\multirow{2}{*}{\bf Model} & \multicolumn{4}{c|}{\bf WN18RR} & \multicolumn{4}{c}{\bf FB15K-237} \\
& \bf MRR & \bf Hit@1 & \bf Hit@3 & \bf Hit@10 & \bf MRR & \bf Hit@1 & \bf Hit@3 & \bf Hit@10 \\ 
\hline
w/o dataset expansion & 68.5 & 61.6 & 72.6 & 80.3 & 34.2 & 25.6 & 36.9 & 51.5 \\ 
w/o balanced multi-task learning & 70.8 & 65.3 & 73.6 & 81.5 & 34.7 & 26.0 & 37.4 & \bf{52.3} \\ 
w/o multi-level shared knowledge & 66.9 & 60.3 & 70.5 & 79.3 & 34.1 & 25.4 & 36.8 & 51.7 \\ 
w/o bi-encoder architecture  & 68.2 & 61.2 & 72.4 & 80.6 & 33.3 & 24.3 & 36.2 & 51.1 \\ 
SKG-KGC & \bf{72.2} & \bf{67.0} & \bf{75.1} & \bf{81.6} & \bf{35.0} & \bf{26.4} & \bf{37.7} & 52.2 \\ 
\hline
\end{tabular}
}
\caption{\label{tab:ablation} The ablation results on the WN18RR and FB15K-237 dataset.}
\end{table*}

{\bf Effect of dataset expansion} Removing dataset expansion causes a significant decrease in our model's performance on the WN18RR and FB15K-237 datasets. Particularly on the WN18RR dataset, which features more textual descriptions, MRR, Hit@1, Hit@3, and Hit@10 metrics all drop by 3.7\%, 5.4\%, 2.5\% and 1.3\%, respectively. This emphasizes the effectiveness of common knowledge within entity sets sharing the same $(h, r)$ or $(r, t)$. Such dataset-level shared knowledge enhances the model's ability to learn common features among interconnected entities.

{\bf Effect of balanced multi-task learning} On the WN18RR and FB15K-237 datasets, when balanced multi-task learning is excluded, the MRR, Hit@1, and Hit@3 of the model show a decrease, but the Hit@10 metric is still comparable. This highlights the advantage of our proposed loss weight allocation scheme for multiple subtasks in multi-task learning. The scheme facilitates more accurate identification of the expected entity from candidate entity sets, despite facing challenges in identifying the top-10 entities.

{\bf Effect of bi-encoder architecture} The removal of the bi-encoder architecture results in a 4\% decrease in MRR on WN18RR, and a 1.7\% decrease on the FB15K-237 dataset. This indicates that it is reasonable for the model to use two independent encoders to encode unknown and known elements separately, thereby avoiding some potential confusion in the single encoder configuration. These findings highlight the effectiveness of the bi-encoder architecture in seamlessly integrating dataset-level and task-level shared knowledge, significantly improving the model's proficiency in knowledge graph completion.

\subsection{Further Exploration of Dataset-level Knowledge}
During dataset expansion, we study how different input texts, the number of sentences, and entity sets affect our model, aiming at further exploration of dataset-level knowledge.

{\bf Experiment 1: Effect of input texts} In this experiment, we assess the impact of different input texts on the model performance. We examine four scenarios: without entity descriptions, without entity names, with both but without text summarization, and with both including text summarization.

\begin{table}[ht]
\centering
\resizebox{\linewidth}{!}{
\begin{tabular}{c|cccc}
\hline
{\bf Input text} & \bf MRR & \bf Hit@1 & \bf Hit@3 & \bf Hit@10 \\ \hline
w/o entity description & 41.5 & 32.7 & 46.0 & 58.1 \\ 
w/o entity name & 65.3 & 57.6 & 70.0 & 79.0 \\ 
w/o text summarization & 70.5 & 64.2 & 74.1 & \bf{81.8} \\ 
SKG-KGC & \bf{72.2} & \bf{67.0} & \bf{75.1} & 81.6  \\ 
\hline
\end{tabular}
}
\caption{\label{tab:inputtext} Performance comparison of different input texts on WN18RR.}
\end{table}

The results in Table \ref{tab:inputtext} indicate that the removal of entity descriptions and names leads to a 30.7\% and 6.9\% decrease in the model's MRR, respectively, underscoring the importance of these features in capturing in-depth semantic relations in the text-based KGC methods. Importantly, entity descriptions contribute significantly to providing an extensive textual context. Furthermore, the application of the TextRank text summarization algorithm yields a 1.7\% increase in MRR, effectively addressing the issue of text redundancy due to an excess of entities.

{\bf Experiment 2: Selection of top-$n$ sentences} During text summarization, we select the top $n$ sentences with the highest TextRank values to serve as concise text, providing the model with the necessary but succinct descriptive information. Accordingly, we explore the impact of the number of sentences on the model's overall performance.

\begin{figure}[ht]
\centering
\includegraphics[scale=0.35]{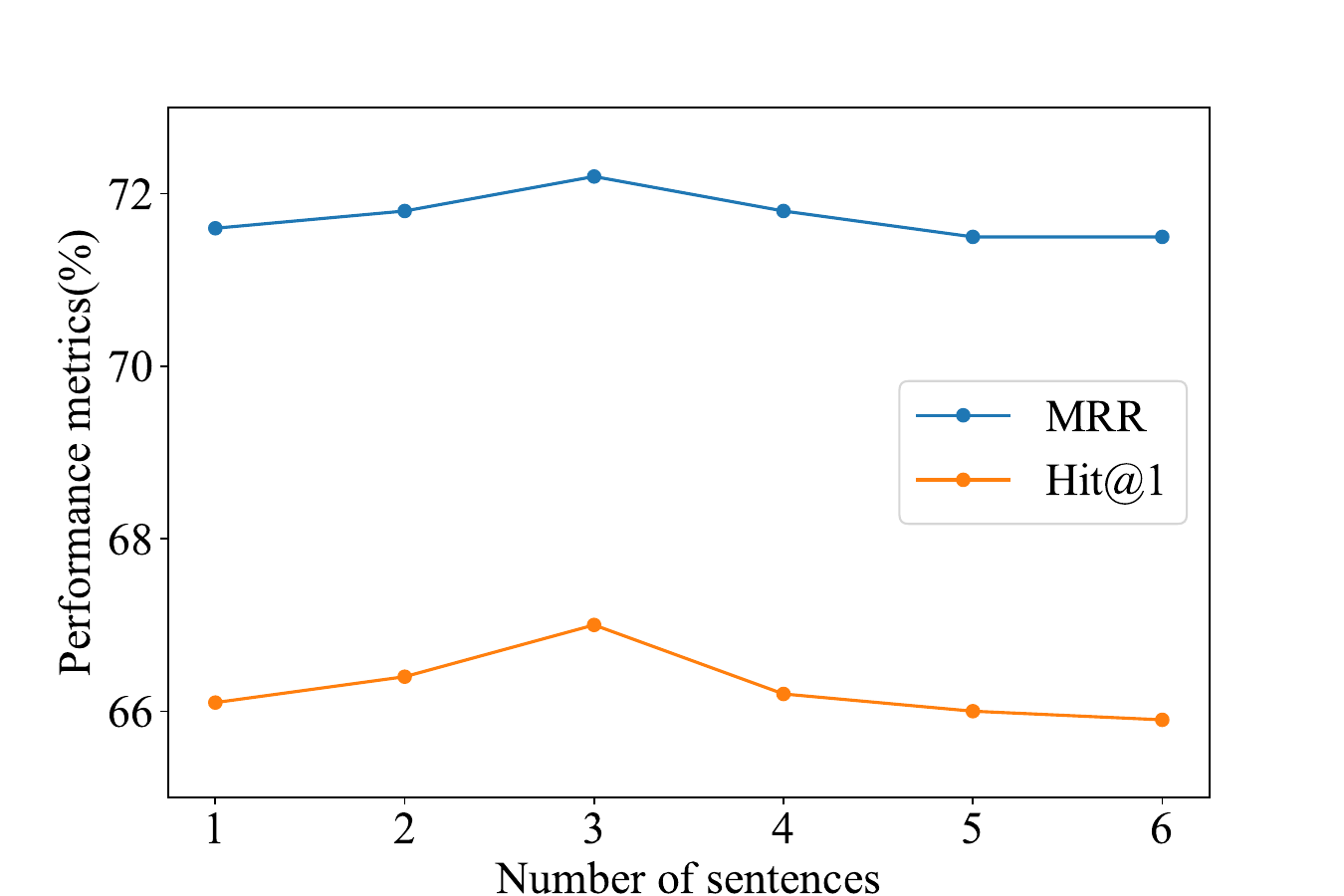}
\caption{\label{fig:3-topn} The impact of the number of sentences on performance metrics.}
\end{figure}

Figure \ref{fig:3-topn} presents the experimental outcomes of selecting the top-$n$ sentences ($n=\{1,2,3,4,5,6\}$) on the WN18RR dataset. When $n=3$, the MRR and Hit@1 metrics of the SKG-KGC model reach their optimal value. When $n$ is less than 3, the model may face challenges in fully comprehending more detailed information regarding the entity context. Conversely, when $n$ increases, the influx of descriptive information might lead to information overload and confusion, making it challenging to identify the more critical contextual information about entities. Consequently, the top three sentences  are ultimately selected as the summarized text. 

{\bf Experiment 3: Effect of entity sets} We compare SKG-KGC with its two variants that remove head entity sets $(\{h_0,h_1,...,h_j\},r,t)$ and tail entity sets $(h,r,\{t_0,t_1 ,...,t_i\})$ on the WN18RR dataset.

\begin{table}[ht]
\centering
\resizebox{\linewidth}{!}{
\begin{tabular}{p{1.8cm}c|cccc}
\hline
{\bf Model} & {\bf Subtask} & \bf MRR & \bf Hit@1 & \bf Hit@3 & \bf Hit@10 \\ \hline
\multirow{3}{1.8cm} {w/o head entity sets} & $(?,r,t)$ & 66.5  & 59.4 & \bf 70.6 & \bf 79.6 \\ 
& $(h,r,?)$ & 73.3 & 65.6 & 78.4  & \bf 87.1 \\ 
& Average & 69.9 &  62.5 & 74.5  & \bf 83.3 \\ \hline
\multirow{3}{1.8cm} {w/o tail entity sets} & $(?,r,t)$ & 66.4 & 60.1 & 70.1 & 77.7   \\ 
& $(h,r,?)$ & 73.8 & 66.6 & 78.6 & 86.5  \\ 
& Average & 70.1 & 63.3  & 74.3  & 82.1 \\ \hline
\multirow{3}{1.8cm} {SKG-KGC} & $(?,r,t)$ & \bf 67.5 & \bf 62.4 & 70.4 & 76.5   \\ 
& $(h,r,?)$ & \bf 76.9 & \bf 71.6 & \bf 79.8 & 86.7   \\ 
& Average & \bf 72.2 & \bf 67.0 & \bf 75.1 & 81.6 \\ \hline
\end{tabular}
}
\caption{\label{tab:entitysets} Effectiveness of different entity sets on WN18RR. $(?,r,t)$ and $(h,r,?)$ denote head entity and tail entity prediction respectively.}
\end{table}

Table \ref{tab:entitysets} shows the effect of such exclusions on the model's performance in predicting head and tail entities. Removing head entity sets significantly reduces the performance of tail entity prediction (MRR decreases by 3.6\%), while removing tail entity sets only slightly affects head entity prediction (MRR decreases by 1.1\%). The overall performance in entity prediction benefits from shared knowledge across all dataset levels, notably for the tail entity prediction task.  We attribute this observed phenomenon to the proportion of triples sharing the same $(r,t)$ or $(h,r)$, as depicted in Figure \ref{fig:1a-share}. The WN18RR dataset contains more triples with the same $(r,t)$, thereby providing a wealth of knowledge about head entity sets and resulting in a more significant enhancement in the tail entity prediction task.

\subsection{Further Exploration of Balanced Multi-Task Learning}
We analyze SKG-KGC alongside weight-unadjusted methods from two perspectives: different datasets and different subtasks of the WN18RR dataset.

{\bf Experiment 1: Performance of balanced multi-task learning on different datasets} As shown in Table \ref{tab:ablation}, balanced multi-task learning works well on WN18RR, but shows only slight improvements on the larger FB15k-237 dataset. We attribute this performance to two main factors. First, the scheme is designed for addressing the issue of imbalanced loss weights among tasks, so it works well when task differences are significant. As shown in Figure \ref{fig:1a-share}, the proportion of triples sharing the same (r, t) or (h, r) is 84.3\% and 74.8\% on the FB15K-237 dataset, respectively, which means less task disparity compared to the 24.8\% on WN18RR.  Secondly, our scheme dynamically updates the loss weights of all tasks after each iteration. Due to computational resource limitations, the number of iterations performed on larger datasets is reduced, leading to less pronounced changes in task weights. Therefore, our proposed scheme performs better when applied to smaller datasets and more diverse tasks.

{\bf Experiment 2: Performance of balanced multi-task learning on different subtasks} Furthermore, Table \ref{tab:task-subtask} provides detailed results on the WN18RR dataset, including head entity and tail entity prediction outcomes. 

\begin{table}[ht]
\centering
\resizebox{\linewidth}{!}{
\begin{tabular}{cc|cccc}
\hline
{\bf Model} & {\bf Subtask} & \bf MRR & \bf Hit@1 & \bf Hit@3 & \bf Hit@10 \\
\hline
\multirow{3}{*} {w/o weight} & {$(?,r,t)$} & 66.3 & 60.9 & 68.4 & 76.8 \\ 
& {$(h,r,?)$} & 75.3 & 69.6 & 78.8 & 86.1 \\ 
& Average & 70.8 & 65.3 & 73.6 & 81.5 \\ \hline
\multirow{3}{*}{SKG-KGC} & {$(?,r,t)$} & 67.5 & 62.4 & 70.4 & 76.5 \\ 
& {$(h,r,?)$} & 76.9 & 71.6 & 79.8 & 86.7 \\ 
& Average & \bf 72.2 & \bf 67.0 & \bf 75.1 & \bf{81.6} \\ \hline
\end{tabular}
}
\caption{\label{tab:task-subtask} Performance of balanced multi-task learning on different subtasks of WN18RR.}
\end{table}

Notably, tail entity prediction consistently outperforms head entity prediction. We attribute this to the smaller average number of entities connected in one-to-many relations. Moreover, Figure \ref{fig:1b-balance} underscores the imbalanced distribution of head entities and tail entities. While our proposed SKG-KGC improves experimental performance through a designed loss weight allocation scheme in multi-task learning, the challenge of significant performance differences between head entities and tail entities persists.

\subsection{Parameter Analysis in Bi-Encoder Architecture}
In our experiment, we utilize two types of encoders: BERT-base-uncased with 110M parameters and BERT-large-uncased with 340M parameters, to assess their performance on WN18RR. Each model is evaluated in two configurations: single-encoder uses one encoder for all elements, while bi-encoder uses two separate encoders for known and unknown elements.

\begin{table}[ht]
\centering
\resizebox{\linewidth}{!}{
\begin{tabular}{cc|cccc}
\hline
{\bf Encoder} & {\bf \#num} & \bf MRR & \bf Hit@1 & \bf Hit@3 & \bf Hit@10 \\ \hline
\multirow{2}{*} {BERT-base (110M)} & one & 68.2 & 61.2 & 72.4 & 80.6 \\ 
& two & \bf{72.2} & \bf{67.0} & 75.1 & 81.6 \\ 
\hline
\multirow{2}{*} {BERT-large (340M)} & one & 69.6 & 62.3 & 74.1 & 82.7 \\ 
& two & 70.8 & 63.8 & \bf{75.2} & \bf{83.2}  \\ 
\hline
\end{tabular}
}
\caption{\label{tab:parameter} Performance comparison of encoders with varying parameter volumes on WN18RR.}
\end{table}

As shown in Table \ref{tab:parameter}, within a bi-encoder architecture, the Hit@10 metric improves with the substantial increase in parameter volume of BERT-large. However, the more critical MRR and Hit@1 metrics decline significantly by 1.4\% and 3.2\%, respectively, potentially due to the curse of dimensionality and overfitting. This observation indicates that an increase in parameter volume does not necessarily lead to an overall improvement in model performance, as supported by previous research \cite{tian2022mitkgc}.

Furthermore, when comparing the performance between single and bi-encoder configurations, it is evident that the bi-encoder consistently outperforms the single encoder configuration. We speculate this could be attributed to the bi-encoder's explicit differentiation between the embeddings of known and unknown elements in the triplets, thereby avoiding potential confusion in the single encoder configuration. Hence, when selecting encoders and architectures for similar tasks, priority should be given to the selection of architecture rather than simply increasing the model's parameter volume.

\subsection{Efficiency Analysis}

Since our model and MTL-KGC \cite{kim2020mtlkgc} both engage in multi-task learning in KGC, we employ the same task settings and encoders for comparative analysis. Table \ref{tab:time} reports the approximate time cost for training and inference.

\begin{table}[ht]
\centering
\resizebox{\linewidth}{!}{
\begin{tabular}{c|ccc|ccc} 
\hline
\multirow{2}{*} {\bf Time} & \multicolumn{3}{c|}{\bf WN18RR} & \multicolumn{3}{c}{\bf FB15K237} \\
& \bf T/EP & \bf Train & \bf Inf & \bf T/EP & \bf Train & \bf Inf  \\ \hline
MTL-KGC & 2.6h & 7.8h & 60h & 6.9h & 20.8h & 491h \\
Our model & 5.5m & 2.5h & 2.8m & 16.7m & 2.7h & 10m \\ \hline
\end{tabular}}
\caption{\label{tab:time} Comparison of time cost between our model and MTL-KGC. The terms "T/Ep", "Train", and "Inf" denote the training time per epoch, the training time until convergence, and inference time, respectively.}
\end{table}

Compared with MTL-KGC, SKG-KGC demonstrates superior speed in training and test datasets. This efficiency improvement emerges from  our model's use of independent candidate entity encoders for calculating entity rankings, similar to the approaches employed by StAR \cite{wang2021star} and SimKGC \cite{wang2022simkgc}. While not pioneering fast inference, our proposed model achieves a trade-off between efficiency and effectiveness, with a focus on improving the latter. Notably, our model surpasses MTL-KGC in both training and inference speed, aligning with the theoretical analysis outlined in section \ref{sec:Inference}.

\subsection{Case study}
To conduct a qualitative analysis of the multi-level shared knowledge, we show the top two entities as ranked by SKG-KGC, SKG-KGC without shared knowledge, and the most competitive baseline SimKGC in Table \ref{tab:case}. 

\begin{table}[ht]
\centering
\resizebox{\linewidth}{!}{
\begin{tabular}{l} 
\hline
\rowcolor{gray!30}
{\bf Case 1: Input (head and relation)}:\\ 
\rowcolor{gray!30}
\quad $h$: Cleopatra [is a 1963 British-American-Swiss epic drama film...] \\ 
\rowcolor{gray!30}
\quad $r$: /film/film/featured\_film\_locations \\ \hline
Prediction (SKG-KGC):\\
\quad P1*: Rome [Located in the foothills of the Appalachian Mountains...]\\
\quad P2*: City of London [is a city within London...] \\ \hline
Prediction (SKG-KGC w/o \textit{Shared Knowledge}):\\
\quad P1: Zurich [is the largest city in Switzerland...]\\
\quad P2*: Rome [Located in the foothills of the Appalachian Mountains...] \\ \hline
Prediction (SimKGC):\\
\quad P1*: Rome [Located in the foothills of the Appalachian Mountains...] \\ 
\quad P2: Zurich [is the largest city in Switzerland...]\\
\hline
\rowcolor{gray!30}
{\bf Case 2: Input (relation and tail)}:\\ 
\rowcolor{gray!30}
\quad $r$: /location/location/contains \\ 
\rowcolor{gray!30}
\quad $t$: Curtis Institute of Music [is a conservatory in Philadelphia...] \\ \hline
Prediction (SKG-KGC):\\
\quad P1*: United States of America [commonly referred to as the United States...] \\
\quad P2: Pittsburgh, PA Metropolitan Statistical Area [is the largest population center...] \\ \hline
Prediction (SKG-KGC w/o \textit{Shared Knowledge}):\\
\quad P1: Pittsburgh, PA Metropolitan Statistical Area [is the largest population center...]\\
\quad P2*: United States of America [commonly referred to as the United States...] \\ \hline
Prediction (SimKGC):\\
\quad P1: Allentown [is a city located in Lehigh County...]\\
\quad P2*: United States of America [commonly referred to as the United States...] \\ \hline
\rowcolor{gray!30}
{\bf Case 3: Input (head and relation)}:\\ 
\rowcolor{gray!30}
\quad $h$: Flo Rida [Tramar Lacel Dillard, better known by...] \\ 
\rowcolor{gray!30}
\quad $r$: /people/person/profession \\ \hline
Wrong Prediction (All models):\\
\quad P1: Record producer-GB [is an individual working within the music industry...] \\
\quad P2: Music executive-GB [is a person within a record label...] \\ \hline
\end{tabular}}
\caption{\label{tab:case} Case study on the FB15K-237 dataset. [*] indicates ground-truth entity. Texts in brackets represent the textual description for entities.}
\end{table}

In the first case, SKG-KGC correctly predicts the entity "\textit{Rome}" and also unexpectedly predicts "\textit{City of London}", possibly due to the influence of "\textit{London}" in the shared tail entity sets. In the second case, SKG-KGC correctly identifies "\textit{United States of America}" by utilizing shared knowledge from candidate entity sets, while other models fail due to an overemphasis on textual similarity between "\textit{Pittsburgh}" / "\textit{Allentown}" and "\textit{Philadelphia}". However, in the third case, the training set only reveals that \textit{Flo Rida}'s profession is that of an actor and songwriter, and the correct tail entity should be \textit{Artist-GB}. All three models predict incorrectly due to the presence of the word "\textit{song}" in \textit{Flo Rida}'s description. Thus, we suspect that text-based methods may excessively focus on certain text descriptions of the entities themselves and overlook structural information in the knowledge graph.

These results highlight that SKG-KGC can mitigate the over-reliance on semantic similarity as compared to previous methods, and effectively improve the ability to identify correct entities from similar candidate entities. Furthermore, these insights prove valuable for considering both textual and structural information in KGC.

\section{Conclusion}
In this paper, we introduce a multi-level shared knowledge guided method for efficient knowledge graph completion. Our approach effectively addresses the challenges of inadequate knowledge learning and imbalanced subtasks in multi-task learning. Through extensive experiments on benchmark datasets, we demonstrate that SKG-KGC consistently outperforms competitive baseline models, particularly excelling on WN18RR with its extensive entity descriptions. These findings provide new insights for multi-task learning and other tasks related to knowledge graphs. In future research, we aim to explore the integration of text-based methods with graph embeddings to extract the semantic and structural information in knowledge graphs.

\bibliography{tacl2021}
\bibliographystyle{acl_natbib}








  

\end{document}